\documentclass[twocolumn,noshowpacs,preprintnumbers,amsmath,amssymb]{revtex4}
\usepackage{graphicx}% Include figure files
\usepackage{dcolumn}% Align table columns on decimal point
\usepackage{bm}% bold math

\newcommand{\beq}{\begin{equation}}
\newcommand{\eeq}{\end{equation}}
\def\be {\begin{equation}}
\def\ee {\end{equation}}
\def\bs#1\es{\begin{split}#1\end{split}}
\def\ba#1\ea{\begin{align}#1\end{align}}
\def\baed#1\eaed{\begin{aligned}#1\end{aligned}}
\def\bged#1\eged{\begin{gathered}#1\end{gathered}}
\def\bea{\begin{eqnarray}}
\def\eea{\end{eqnarray}}

\usepackage{xcolor}
\usepackage[colorlinks = true,
            linkcolor = blue,
            urlcolor  = blue,
            citecolor = blue,
            anchorcolor = blue]{hyperref}

\newcommand{\cref}{{\bf [check ref]}}

\usepackage{amsfonts}
\usepackage{booktabs}
\usepackage{siunitx}

\begin{document}
\preprint{ } % \tiny matthias.weissenbacher@ipmu.jp}
\title{ Temporally Folded Convolutional Neural Networks for Sequence Forecasting
}% Force line breaks with
\author{ Matthias Weissenbacher }
\affiliation{%
 {\small Kavli IPMU (WPI), University of Tokyo, Japan} % {\tiny, Kashiwa-no-ha 5-1-5, 277-8583, Japan}
}
\email{  matthias.weissenbacher@ipmu.jp}
%\emailAdd{matthias.weissenbacher@ipmu.jp}

\begin{abstract}
In this work we  propose a novel  approach to utilize convolutional  neural networks for time series forecasting. 
The time direction of the sequential data with spatial dimensions $D=1,2$  is considered democratically as the input of a spatiotemporal $(D+1)$-dimensional convolutional neural network. Latter then reduces the data stream from $D +1 \to D$ dimensions followed by an incriminator cell which uses this information to forecast the subsequent time step. We empirically compare this strategy  to convolutional LSTM's and LSTM's on their performance on the sequential MNIST and  the JSB chorals dataset, respectively. We conclude that temporally folded convolutional neural networks (TFC's)   may outperform  the  conventional recurrent strategies.

\end{abstract}
%\pacs{} PACS, the Physics and Astronomy
\maketitle

%%%%%%%%%%%%%%%%%%%%%%%%%%%%%%%%%%%%%%%
\section{Introduction}
%%%%%%%%%%%%%%%%%%%%%%%%%%%%%%%%%%%%%%%

Time series forecasting admits a wide range of applications from signal processing, pattern recognition and weather forecasting to  mathematical finance, to name only a few.  Machine learning techniques for time-series forecasting have been widely studied \cite{Goodfellow-et-al-2016,NgAndrew}.

The traditional recurrent approaches towards sequence modeling tasks \cite{Goodfellow-et-al-2016,NgAndrew} have been recently challenged by  convolutional network architectures \cite{oord2016wavenet,kalchbrenner2016neural,gehring2017convolutional,dauphin2016language}. Latter  compete in the categories speed and precision and regularly outperform conventional recurrent approaches such as LSTM's, GRU's or RNN's \cite{chung2014empirical,pascanu2013construct,Jozefowicz:2015:EER:3045118.3045367,zhang2016architectural,Greff_2017,hewamalage2019recurrent}.  
In particular, those convolutional  architectures  may overcome the deficiencies of recurrent networks to handle long and multi-scale sequences with increased  receptive fields \cite{oord2016wavenet, lea2016temporal,2018arXiv180301271B}.
%Convolutional networks applied to sequences reach back to
For time sequences  of images convolutional LSTM's aim to combine the best of both worlds \cite{NIPS2015_5955,wiewel2018latentspace}. 

\begin{figure}[!ht]
    % \centering 
    \vspace{-0.1cm}
\begin{center}
%\hspace{-0.4 cm}
\includegraphics[height=7.8cm]{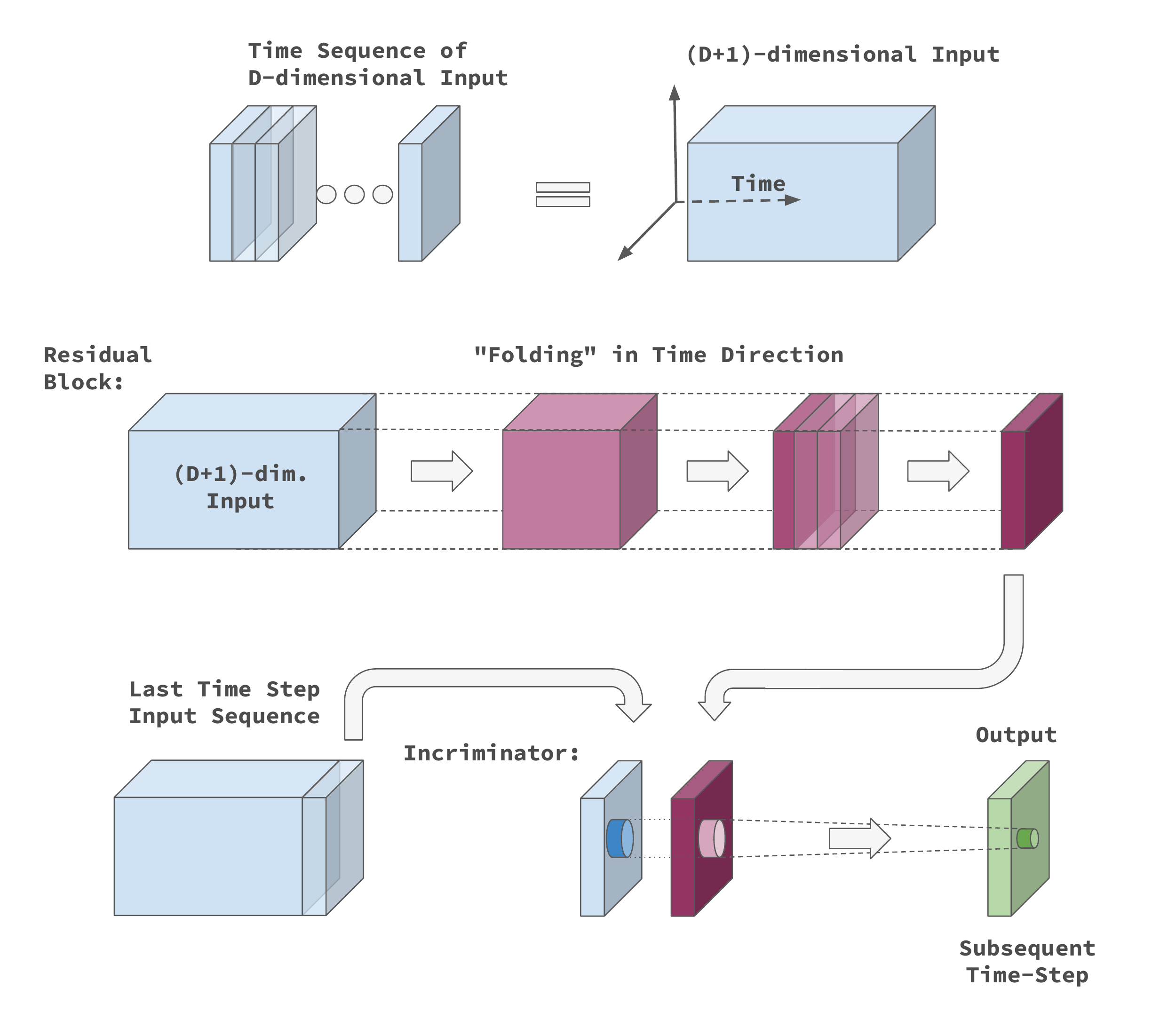}
\end{center}\vspace{-0.2cm}
     \caption{Architecture of a temporally folded convolutional network (TFC). The first row depicts the rearrangement of the D-dimensional "spatial"  time sequence data into a (D+1)-dimensional array. Secondly, the action of the residual block  on the  (D+1)-dim.\;spatiotemporal input is shown. The arrows  illustrate the action of individual residual cells with strides bigger than one in the time direction whilst the spatial format remains unchanged. We highlight the increase in features from left to right by  an increasingly intense color scheme. The incriminator cell then connects local neighborhoods around any point in the two  spatial data slices to predict the D-dimensional output, i.e.\,the subsequent time-step.  }\label{fig:TFC}
   \end{figure}

In this work we present a novel approach to  utilize convolutional  neural networks for image sequence as well as general sequence forecasting  tasks. 
In contrast to the recent serge in causal "dilated" convolutional networks \cite{oord2016wavenet,kalchbrenner2016neural,gehring2017convolutional,dauphin2016language, lea2016temporal,2018arXiv180301271B,aksan2019stcn,singh2018recurrent}  our approach is closer in spirit to non-casual architectures \cite{long2014fully,10.1007/978-3-642-15567-3_11,aneja2017convolutional,choy20194d}.  
However, our architecture  distinguishes itself by its composite design for time series forecasting,  see fig.\,\eqref{fig:TFC}.
The time and spatial directions of  sequential data  is considered democratically as non-causal input  of the convolutional part of the network. Referred to as residual block it is a collection of residual convolutional cells \cite{he2015deep}. An  "incriminator" cell then uses this output to make a strictly causal prediction for the next time step. More precisely, a temporally folded convolutional network (TFC) is the combination of a residual block which reduces the data stream from $D +1 \to D$ dimensions by "folding'" the time direction. Followed by  another cell which uses this information to forecast the subsequent time step. We refer to latter as the incriminator cell.  Note that this architecture establishes  correlations over the entire time span of the input data while maintaining small kernel sizes.
\newline
Moreover, we argue that the TFC architecture can be applied for image classification tasks by  subjecting the incriminator cell to a minor adjustment. This may constitute an interesting direction to gain a better understanding of how deep neural networks generalize. 
\vspace{-0.1cm}

%Note that the sequential MNIST image forecasting may be achieved with ease by any human. In contrast to continue the JSB chorals time-series  i.e. to predict which keys are played  next on the piano may not be achieved unless one is a trained musician. In this light it is every encouraging that the TFC performs well on the JSB chorals dataset. 

This letter is structured as follows. The theoretical concept and architecture of temporally folded convolutional networks  are introduced in section \ref{sec:TFC}.
In section \ref{sec:MNIST} we then implement several variations of $D=2$ TFC architectures and systematically  compare them to conventional convolutional LSTM's in their performance on the sequential MNIST dataset.  We proceed analogously in section \ref{sec:JSB} where we compare a $D=1$ TFC architecture to a LSTM on the JSB choral dataset \footnote{Note that this  deviates from the standard use of the JSB dataset in literature\ cite{}. }. Moreover, we argue that the TFC architecture  generalizes to be applicable to the CIFAR10 image classification task.  The trained models as well as source code for the networks architecture will be provided \footnote{Provided after peer review. Please visit \href{www.github.com}{Github}. }.

%Let us mention that it would be interesting to compare the TFC architecture to the performance of dilated convolutional networks for time-series forecast \cite{}.

%%%%%%%%%%%%%%%%%%%%%%%%%%%%%%%%%%%%%%%
\section{ Temporally folded convolutional networks}\label{sec:TFC}
%%%%%%%%%%%%%%%%%%%%%%%%%%%%%%%%%%%%%%%

The essence of this section is to describe a different angle on time series forecast utilizing deep residual convolutional neural networks. Let us give the basic definition of the TFC network architecture as
\beq  
\mathcal{N}_{TFC} =   \mathcal{I} \circ \mathcal{R} \;\, ,
\eeq
where $ \mathcal{I} $ is the incriminator cell and  $\mathcal{R} $ the residual block. The network acts on the input sequence according to fig.\,\eqref{fig:TFC}. The simplest example for a concrete implementation would be to take $\mathcal{R} $  to be a $D+1$-dimensional  convolutional layer with strides equal to the D-dimensional sequence  length. This then produces a D-dimensional data slice.  The simplest example for   $\mathcal{I} $ is a sequence of a fully connected layer with a preceding flattening and concatenation layer and a final reshaping to the  D-dimensional output shape. 

{\bf Comparison to dilated convolutional nets.}
Let us next provide a qualitative comparison of the TFC structure relative to dilated convolutional networks 
\cite{oord2016wavenet,kalchbrenner2016neural,gehring2017convolutional,dauphin2016language, lea2016temporal,2018arXiv180301271B,aksan2019stcn,singh2018recurrent}  for time series modeling. 
\begin{figure}[!ht]
    % \centering 
    \vspace{-0.3cm}
\begin{center}
%\hspace{-0.4 cm}
\includegraphics[height=3.05cm]{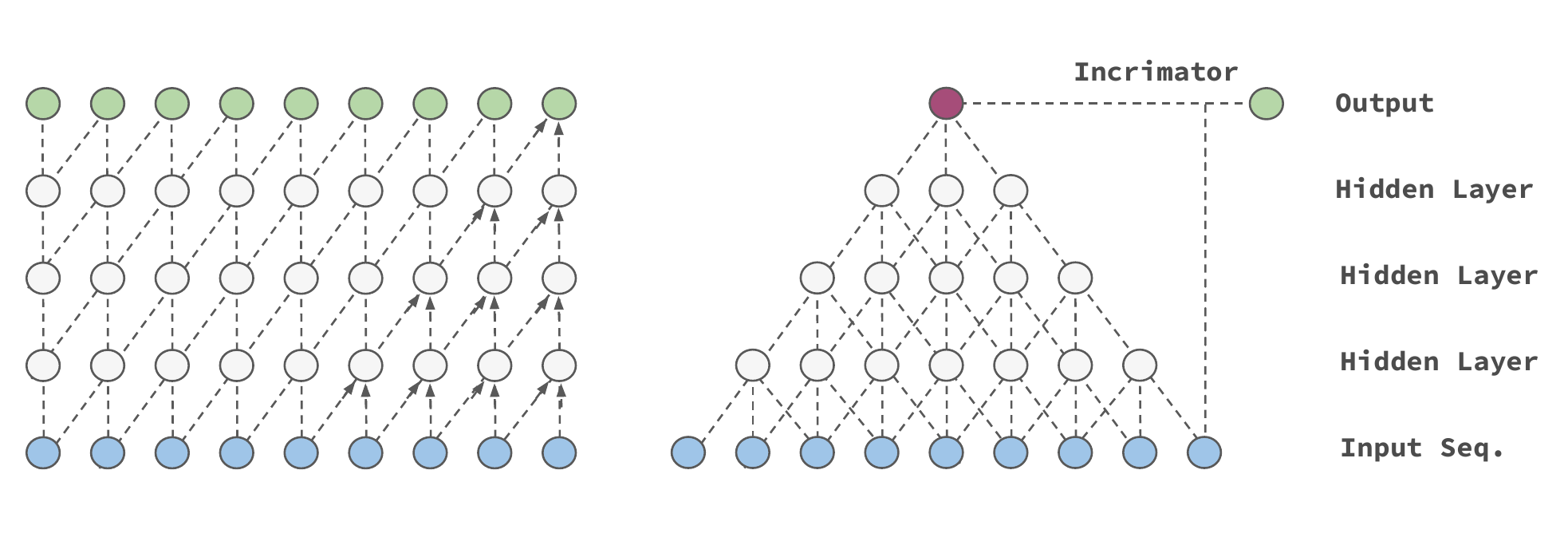}
\end{center}\vspace{-0.4cm}
     \caption{ Visualization of a stack of causal dilated convolutional layers (left) versus the TFC's non-causal convolutional (residual) block and causal incriminator (right). The last layer on the top right is the incriminator cell. The forward time direction is from left to right in each depiction, respectively.}\label{fig:TFCdil}
   \end{figure}\vspace{-0.1cm}
Latter usually produces  an output sequence of length equal to the input sequence whilst here we focus on forecasting one time step. However, the main intrinsic differences can be inferred from  fig.\eqref{fig:TFCdil}. Firstly, dilated convolutional networks  exclusively use preceding time-step information in each hidden layer (causal). Whilst the TFC architecture's convolutional part  i.e.\,the residual block depicted as the pyramid in fig.\eqref{fig:TFCdil} is non-casual, thus future time steps get processed equally to past ones. However, the incriminator cell makes the TFC network strictly casual which is crucial for its application to forecasting tasks. 
Secondly, the TFC architecture democratically uses spatial and time information as input of the convolutional layers.
The time direction of  the sequential D-dimensional spatial data depicted as blue disks in fig.\eqref{fig:TFCdil}  is considered democratically as $D+1$-dimensional spatiotemporal  input see fig.\,\eqref{fig:TFC}.  Thus this architecture distinguishes itself uniquely  from others in the literature.  

 Note that the dimensionality of the data is to be distinguished from its depth along each dimensions, e.g. D=2 corresponds to image data, the depth refers to the pixel width and height of the latter. In this work we empirically focus on  $D=1,2$, however the networks definition holds for arbitrary dimensions.

{\bf Formal definition.} Let us next give a more formal definition of a TFC.
We denote $x_t^{\tiny(m)}$ the D-dimensional data at time $t =0,1,2\dots$ thus one may express the input time sequence as $\{x^{\tiny(m)}_0,\dots,x^{\tiny(m)}_t\}$, where the  superscript denotes the number of features. The residual block with $n$-features is a map
\beq\label{resBlock}
\mathcal{R}_n : \{x_0^{\tiny(m)},\dots, x_t^{\tiny(m)}\} \rightarrow \tilde y^{\tiny(n)} \;\;.
\eeq
Taking e.g.  the  case of a of  color image as input sequence $m=3$  counting the RGB color spectrum, which is then mapped by eq.\eqref{resBlock} to an image of equal dimensions $ \tilde y^{\tiny(n)}$ with n-features. The incriminator  with r internal features is a map 
\beq\label{IncCell}
\mathcal{I}_r :   \{ \,  x_t^{\tiny(m)}  ,  \, \tilde y^{\tiny(n)}\} \rightarrow y_{t+1}^{\tiny(m) } \;\; .
\eeq
 Note that  the dimensionality of the  in- and output  features  of eq.\eqref{IncCell} are entirely fixed. By combining eq.\eqref{resBlock} and  eq.\eqref{IncCell} one infers the architecture of the TFC network to be
 \beq \label{TFC_def}
 \mathcal{N}^{\tiny(n,r)}_{TFC} =   \mathcal{I}_r \circ \mathcal{R}_n : \{x_0^{\tiny(m)},\dots, x_t^{\tiny(m)}\} \rightarrow y_{t+ 1}^{\tiny(m)} \;\; .
 \eeq
Note that for notational simplicity  in  eq.\eqref{TFC_def}  we omit stressing that the incriminator cell takes as well the last  time-step  of the input sequence as  argument eq.\eqref{IncCell} to predict the subsequent time step $y_{t+ 1}^{\tiny(m)}$ \footnote{We use the letter $y$ for the D-dimensional processed data. Also let us note that we employ abuse of notation for the symbol $\circ$  throughout the work. However, we explain its precise functionality in the text. Thus if not specified else-wise for two functions $f,g$ it denotes $f \circ g = f(g)$.}.

 {\bf Residual cell architecture.}  A few comments in order.  
The dimensional reduction $D +1 \to D$ by the residual block $\mathcal{R}_n$ is performed by step-wise "folding" the time direction i.e.\,the pooling or strides are repeatedly applied  solely in the time direction. Thus it captures correlations lengths across the entire time span of the input stream while admitting small kernel-sizes. With the residual block composed of $k$-residual cells $R^{(n_k)}_{s_k}$ where $n_k$ are the features of the k-the cell and $s_k$ the strides along the time direction. Residual cells  are composed of $(D+1)$-dimensional convolutional layers \cite{he2015deep}. Let us next state the above in a more formal manner 
\beq\label{residual_Block}
\mathcal{R}_n :=    \,  \big( \stackrel{\text{\tiny fully}}{\text{\tiny connected}}\big)^{\tiny(n)}   \circ   R^{\tiny (n_k)}_{s_k} \circ  \dots   \circ  R^{\tiny (n_2)}_{s_2}  \circ  R^{\tiny (n_1)}_{s_1} \;\; .
\eeq
The strides are constrained such that it reduces the data stream in the time direction to size one, at  latest at the k-th step.
Furthermore, note that it has proven useful to impose the ordering $m < n_1 < n_2 < \dots < n_k $.

Lastly, note that although the convolutional residual  block \eqref{residual_Block} is non-causal the TFC architecture is.  In fact its strength  is likely to origin from the democratic information processing of the spatiotemporal data-stream.  
Let us  emphasize that  this is not a truly new architecture,  but more a promising strategy composed of different known concepts.  

Note that below we propose an explicit implementation of an incriminator cell, which we then utilize in the remainder of this work. However, let us stress that its  explicit structure may be varied for different purposes, and the label "incriminator cell" stands for its specific role inside the TFC.

{\bf Explicit incriminator cell architecture. } Let us next turn to the explicit  architecture of the incriminator cell we use in the remainder of this work namely $\mathcal{I}_{r,t,N}$ with the three hyper-parameters $r,t$ and $N$ . It can simply be thought of as a "locally full connected" layer of $\tilde y^{\tiny (n)}$ and $x_{t}^{\tiny (m)}$ in regard to their spatial D-dimensional directions.  More precisely, for every neighborhood of  size $r^D$ around any point in $x_{t}^{\tiny (m)}$ it takes the same point in $\tilde y^{\tiny (n)}$ and its neighborhood of size $t^D$ and connects them by a linear weight matrix multiplication to a vector of dimensions $N$. Thus the matrix is of size $\mathbb{R}^{r^D} \times \mathbb{R}^{t^D} \times \mathbb{R}^N$. Secondly, another weight matrix of dimension $\mathbb{R}^N \times \mathbb{R}^m$ is applied to reproduce the correct  number of output features $y_{t+1}^{\tiny (m)}$.  We omit technical details such as e.g.\,biases here and refer the reader to the appendix \ref{app:DetailsArch}  for details. As the same matrices are applied for every point in the D-dimensional spatial directions of the data it is like locally fully connected layer. 
This concept is useful as it reduced the  number of weights drastically in contrast to a fully connected layer in between $\tilde y^{\tiny (n)}$ and $x_{t}^{\tiny (m)}$. It however, implies that to work in practice the spatial correlation length across one time step is smaller than $r^D$. In other words only information inside this surrounding box  around every point is considered to derive the spatial information at the subsequent time step. In e.g. the simple case of a data moving at constant velocity it would need to have  speed smaller than $r / \Delta t$  to be captured directly by this layer. Where $\Delta t$ denotes the time interval  in between $x_{t}^{\tiny (m)}$ and $x_{t+1}^{\tiny (m)}$ \footnote{We use abuse of notation to highlight the connection to constant velocity. Note that the time interval in between $x_{t}^{\tiny (m)}$ and $x_{t+1}^{\tiny (m)}$ per our notation is one. }. 

Secondly, we  introduce a modification of the incriminator cell by adding a D-dimensional locally connected layer $\mathcal{F}_s$  \cite{KerasLocal}, with kernel size $s$   as
\bea\label{incCelladvanced}
 \mathcal{I'}_{r,t,s,N} \big( \{ \,  x_t^{\tiny(m)}  ,  \tilde y^{\tiny(n)}\} \big) :=  \mathcal{I}_{r,t,N}\big(\{ \,  x_t^{\tiny(m)}  ,  \,\mathcal{F}_s \big( \tilde y^{\tiny(n)} \big)\} \big)
\eea
where $ \tilde y^{\tiny(n)}$ is the output of the residual block eq.\eqref{resBlock}. We implicitly define the number of features of the locally connected layer to be $n$  \footnote{The padding is to be such that the format of the D-dimensional data is unchanged. }.

In particular, in section \ref{sec:MNIST} and \ref{sec:JSB} we use  the hyperparamters $t=1$ and $s=1$. As a general note  for $D = 2$  one may choose to generalize the incriminator cells   to allow for different height and width of the selected neighborhood. However, for simplicity we refrain from introducing another hyperparamter in this work.

\textbf{Multi-time step forecast.} Lastly, we consider the case to forecast multiply time steps using the TFC. 
The natural strategy is to train the network $\mathcal{N}^{\tiny(n,r)}_{TFC}$ on forecasting one time-step and then apply serial chaining of latter as
\vspace{-0,3cm}
\bea\label{mulitple_forcast} %\vspace{-0,1cm}
&\stackrel{k-times}{\mathcal{N}^{\tiny(n,r)}_{TFC} \circ .... \circ \mathcal{N}^{\tiny(n,r)}_{TFC} } \;\; :  \;\; \\ \nonumber
& \{x_0^{\tiny(m)},\dots, x_t^{\tiny(m)}\} \;\; \rightarrow \;\;  \{ y_{t+1}^{\tiny(m)},\dots,\, y_{t+ k}^{\tiny(m)} \} \;\; ,
\eea
where the $k^{th}$ TFC unit takes the output of the forgoing unit and the input sequence of latter by omitting its first element. 
Thus e.g.\,for two forecasted times steps one finds
\vspace{-0,3cm}
\bea\label{mulitple_forcast_ex}
\;\;&\;\; \mathcal{N}^{\tiny(n,r)}_{TFC} \circ \mathcal{N}^{\tiny(n,r)}_{TFC} \big(\{x_0^{\tiny(m)},\dots, x_t^{\tiny(m)}\} \big) \;\;  =  \;\; \\\nonumber \vspace{0,3cm}
& = \;\;  \mathcal{N}^{\tiny(n,r)}_{TFC} \big( \{x_1^{\tiny(m)},\dots, x_t^{\tiny(m)}, y_{t+1}^{\tiny(m)} \} \big) =  \{ y_{t+1}^{\tiny(m)},\, y_{t+ 2}^{\tiny(m)} \}\;\; .\vspace{0,2cm}
\eea
Let us stress that this is the standard approach also found  often when employing recurrent networks.
\newline
One may also employ a parallel  training strategy. The difference in contrast to the serial method eq.\eqref{mulitple_forcast} is that the network architecture itself is modified to allow to be trained on multiple time-step forecast predictions. The network architecture is schematically the same as eq.\eqref{mulitple_forcast}  and eq.\eqref{mulitple_forcast_ex} but with the difference that the entire chain is subject to the training process. In other words the  residual block and incriminator cell  pair eq.\eqref{resBlock} and eq.\eqref{IncCell} are applied repeatedly for every additional forecasted time step, i.e.\,the weights are shared among all time steps. %Re-expressing the example \eqref{mulitple_forcast_ex} in this context results in
Note that one could introduce novel weights for the incriminator cell of each forecasted time-step which may further increase the performance of the network.

%%%%%%%%%%%%%%%%%%%%%%%%%%%%%%%%%%%%%%%
\section{ Sequential MNIST dataset}\label{sec:MNIST}
\begin{figure}[!ht]
     \centering
     \begin{tabular}{SSSSS} \toprule
    {$\text{ \tiny \;\;  Truth \;}$} & {$\text{\tiny \;\; \, \; \textbf{TFC-D2-L'} \;}$} &  {$\text{\tiny  ConvLSTM-L} \,\,$}  &   {$\text{ \tiny \;TFC-D2-L  \, \,}$} & {$\text{\tiny\;\;\; \textbf{TFC-D2'}  \; }$} 
\end{tabular}\vspace{-0.1cm}
\begin{center}
\includegraphics[height=10.0cm]{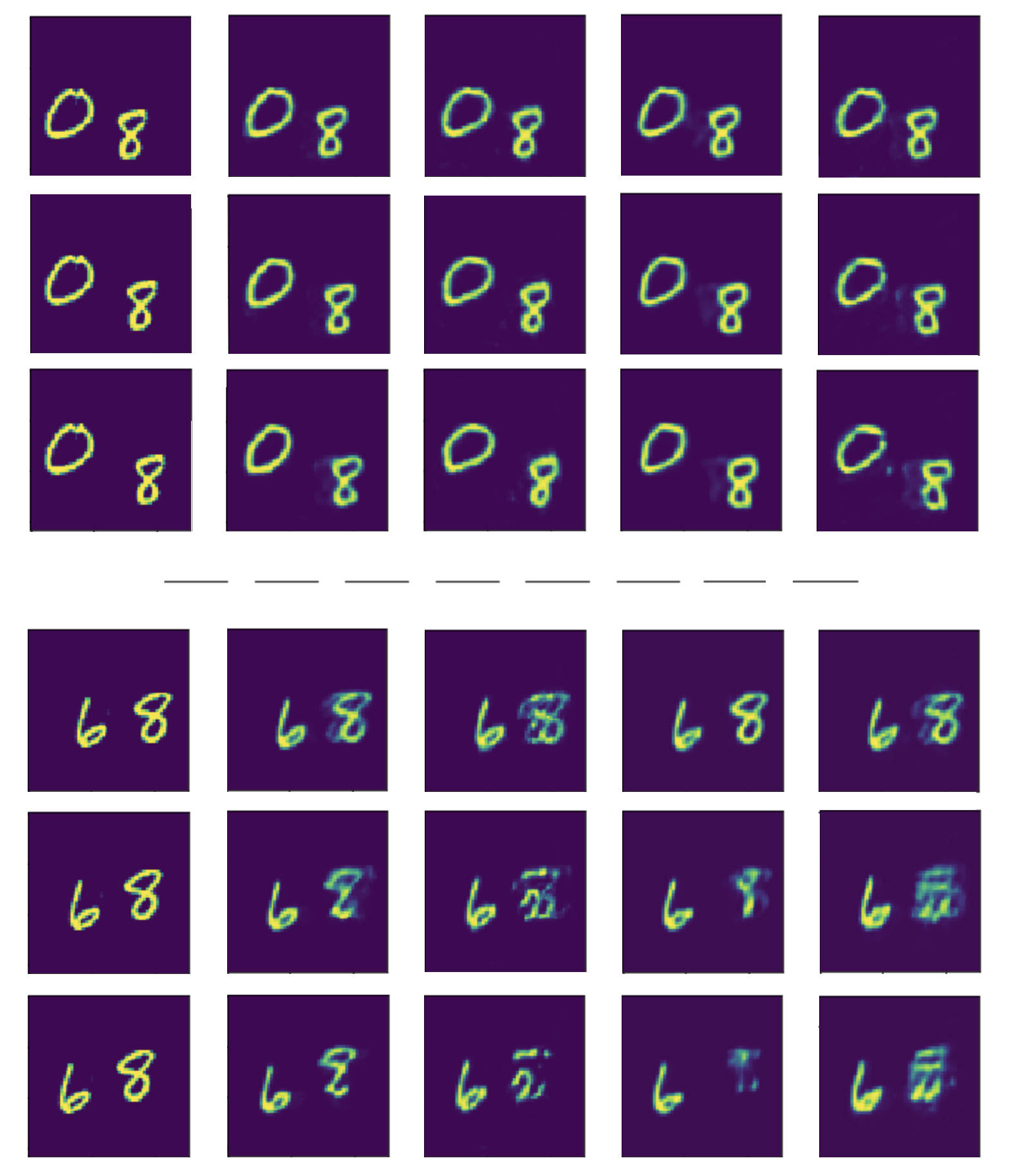}
\end{center}\vspace{-0.1cm}
     \caption{ Sequential MNIST three-step forecast.  The forward time direction is downwards. The figure shows two example data series.  From left to right ground truth, TFC-D2-L', ConvLSTM-L, TFC-D2-L and TFC-D2'.}\label{fig:results_MNIST_Plot}
   \end{figure}

%%%%%%%%%%%%%%%%%%%%%%%%%%%%%%%%%%%%%%%
The sequential or moving MNIST dataset consists of two handwritten digits  moving with constant velocity and random initial position  in a frame of $64 \times 64$ pixels \cite{2015arXiv150204681S}.  The dataset  \footnote{ \href{http://www.cs.toronto.edu/~nitish/unsupervised_video/}{ Link to dataset at http://www.cs.toronto.edu} } consists of   $10.000$  sequences 20 frames long of which we use 10 frames for the input to predict the 11$^\text{th}$ frame and the 11$^\text{th}$-13$^\text{th}$ frame i.e.\,an one-step and a three-step forecast, respectively.  The moving digits are chosen randomly from the MNIST dataset.  
Details of  TFC architecture can be found in the appendix \ref{app:DetailsMNIST}.
We compare its performance to a two-dimensional convolutional LSTM architecture \cite{NIPS2015_5955}. The training of all networks in this section is carried out  over fifty epochs with a batch-size of eighteen. We use the Adam optimizer with learning rate $0.0005$ and the mean squared error (mse) objective function.
\newline
\textbf{Summary results.}  Let us next provide a qualitative overview of the results before turning to the detailed quantitative comparison.  The two  recurrent networks  admit each three  two-dimensional convolutional LSTM  layers, however the ConvLSTM-L admits a larger number of features compared to the ConvLSTM. The TFC architectures  - TFC-D2, TFC-D2'  - admit three and  - TFC-D2-L, TFC-D2-L'  -  five residual cells. The prime denotes the use of the modified incriminator cell \eqref{incCelladvanced}. All networks are designed to forecast one-time step. The three-time step forecast is obtained by serial chaining of the trained network, see eq.\eqref{mulitple_forcast}. The details of the networks and the residual cells are discussed in the appendix \ref{app:DetailsArch}, \ref{app:DetailsMNIST} and \ref{app:JSBDetails}. 
\newline
 We conclude that  the TFC architectures achieve a better performance whilst taking less time to train, see fig.\eqref{fig:results_MNIST}.  The one time step forecast of the  un-primed incriminator cells seems promising. However, the analysis of the three-time step forecast  reveals that those fail to learn that the moving digits bounce off the boundaries, see fig.\,\eqref{fig:results_MNIST_Plot}.
\begin{figure}[!ht]
     \centering\vspace{-0,4 cm}
\begin{tabular}{cccccc} \toprule 
    {$\text{networks}$} & {$\text{ \; weights \; }$} &  {$\stackrel{\text{\; one-step \;}}{\text{mse }}$} &  {$\stackrel{\text{\;three-step\; }}{\text{mse }}$} & {$\stackrel{\text{training-time}}{\text{ {\tiny rel. to TFC-D2}}}$}  \vspace{0.2cm}\\ \midrule \vspace{0.04cm}
      $\text{\small ConvLSTM}$  & $\text{109k}$ & 0.0442  &  0.0729 &  2.1  \\ \vspace{0.04cm}
    $\text{\small ConvLSTM-L}$  & $\text{ 433k}$ &  0.0364  &  0.0622 &     3.3 \\  \vspace{0.04cm} % 0.0367  &  0.0618 
    $\text{\small TFC-D2}$  & $\text{447k}$&  0.0439 &  0.0757  &  1.0 \\ \midrule\vspace{0.04cm}
    $\text{\small  \textbf{ TFC-D2'}}$  & $\text{\textbf{ 742k}}$  & \textbf{  0.0388 }&  \textbf{ 0.0655}  & \textbf{ 1.5 }    \\\vspace{0.04cm}
     $\text{\small TFC-D2-L}$  & $\text{1028k}$  &   0.0366  &  0.0649 &  1.6  \\  \vspace{0.04cm} 
     $\text{\small   \textbf{ TFC-D2-L'} }$  & $\text{\textbf{1272k}}$  & \textbf{ 0.0336}  &  \textbf{0.0584} & \textbf{2.2}  
     \\ \bottomrule
\end{tabular} \vspace{-0,2 cm}
 \caption{Performance comparison in between convolutional LSTM and  TFC architectures. The training time column gives dimensionless factors relative to the TFC-D2 networks $1.9 \, \text{min/epoch}$. All networks architectures are designed to forecast one-time step. The column "three-step" for the other networks makes use of the technique described around eq.\eqref{mulitple_forcast}. 
 %Reference values for the mse compared to the black image  $0.?$ and  to the last image of the input sequence $0.096?$, where we averaged over the test set.
 }   \label{fig:results_MNIST}
\end{figure}

 \vspace{-0,1 cm}

Let us next argue for general features  by pairwise direct comparisons of networks based on  fig.\,\eqref{fig:results_MNIST}  and  fig.\eqref{fig:results_MNIST_Plot}.
\newline
\textbf{ConvLSTM-L vs. TFC-D2-L'.}  Although the TFC-D2-L' network admits a higher number of  total  and trainable weights it takes less time to complete the training process. The mean squared errors for the one-step as well as three step forecast lie significantly below that of the ConvLSTM-L. In particular, the overall shape of the digits is maintained after bouncing off the boundary.
\newline
\textbf{ConvLSTM-L vs. TFC-D2.}  Whilst the TFC-D2 architecture performs equally well on  the one time step compared to the ConvLSTM-L.  The three time-step forecast looses value due to its shortcoming of not being able to predict the digits trajectory after hitting the boundary.
\newline
\textbf{TFC-D2' vs. TFC-D2-L'.} Those two architectures are highlighted in  fig.\eqref{fig:results_MNIST} as those admit the best overall performance. Both  learn all qualitative features of the training  for the one as well as the three-step forecast. The networks are distinguished only by a larger number of residual cells of TFC-D2-L', which results in its better performance.
\newline
Concludingly from comparison of    TFC-D2 to TFC-D2-L  and  TFC-D2' to TFC-D2-L', respectively, one infers that an increase of residual cells
increases the performance of the networks.  

%%%%%%%%%%%%%%%%%%%%%%%%%%%%%%%%%%%%%%%
\section{ JSB chorals  dataset }\label{sec:JSB}
%%%%%%%%%%%%%%%%%%%%%%%%%%%%%%%%%%%%%%%

In this section we aim to show that the TFC architecture may be extended to different tasks by using it to predict future time-steps in the JSB Chorales \cite{boulangerlew2012modeling}  polyphonic music dataset. 
It consists of the entire corpus of 382 four-part harmonized chorales by J. S. Bach. We use 250 chorals for training and test our result on the remaining 132 chorals. 
Each element of the $D=1$-dimensional time sequence is an vector containing $0$'s and  $1$'s that corresponds to the 88 keys on a piano. With integer number 1 indicating that a key  is pressed at any given time.  We then proceed and divide the training and testing set in all possible sequential sequences of length eleven, the first ten sequence elements are used as an input to predict the eleventh element. In total that provides us with a training set of $13,090$ and a test and validation set of $5,809$, latter split in half randomly. For reproducibility matters the performance results  of the networks fig.\,\eqref{fig:results_JSB_table} are provided on the entire  set of the $5,809$ time series.
Let us next turn to the details of the networks. For the TFC we use the analogous architecture as in \ref{sec:MNIST} but for spatial input dimensions D=1, i.e.\,for two-dimensional convolutional layers in the residual cells.  We denote latter by TFC-D1. We use  the ADAM optimizer with learning rate $0.002$.  As the LSTM perform worse on that task at hand we choose its number of weights such that the training time per epoch is comparable to the TFC-D1. Both networks are trained for fifty  epochs with a batchsize of fifty and a mean squared error objective function. Details of the network can be found in the appendix \ref{app:JSBDetails}.
%\vspace{1.2cm}

\begin{figure}[!ht]
     \centering
\begin{center}
\includegraphics[height=8.0cm]{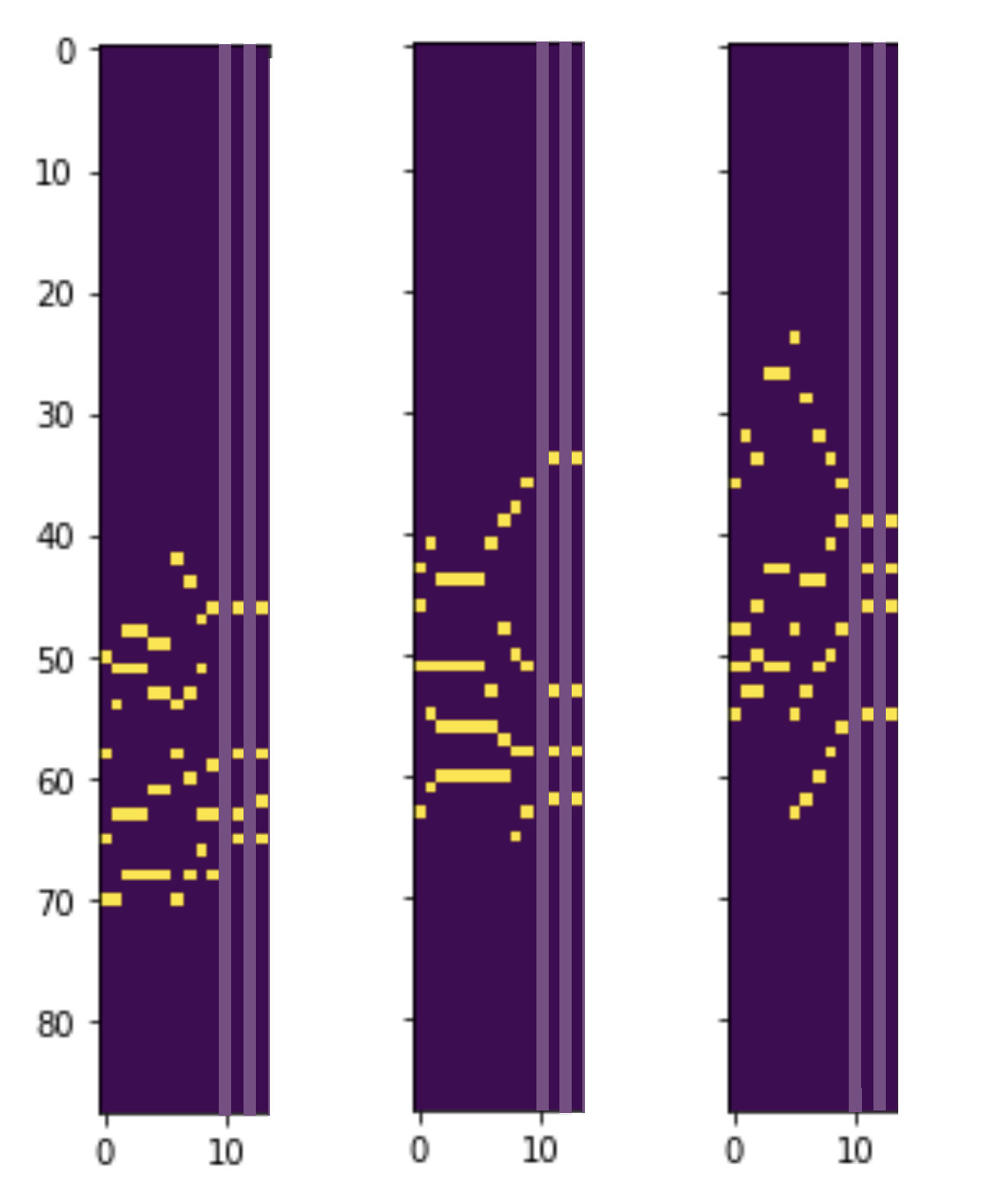}
\end{center}\vspace{-0.1cm}
     \caption{Illustration of the  88 keys of the piano on the vertical axis versus  time-steps  of the JSB chorals datas set on the horizontal axis. The bright yellow color depicts keys which are played at any given time-step. The graph shows three independent input sequences each of length ten followed by their ground truth for the eleventh subsequent time-step, followed by the prediction of the TFC-D1 model, from left to right. The lighter area  serves purely a graphical function.  }\label{fig:results_JSB}
   \end{figure}

\textbf{LSTM vs. TFC-D1.} The TFC architecture outperforms the classical LSTM at the JSB choral time series forecast see fig.\,\eqref{fig:results_JSB_table}. Note that at comparable training time the LSTM admits more than double the number of weights  as the TFC-D1. The custom definition for the accuracy  is \footnote{We define a custom accuracy for the JSB time series forecast as the standard accuracy function give values close to $98\%-100\%$ due to the large  number of keys not played at any given time. }
\beq
acc(T,P) = 1 - \frac{mse(T,P)}{\sum_i T_i + P_i} \; ,\;\; i = 1,\dots,88 \;\;.
\eeq
 where $T,P$ are the test and prediction  for  a single sequence converted to boolean values $\{ 0,1 \}$, respectively. To derive this intuitive boolean quantity we convert the output of the network to interpret values above/below a threshold as keys played/not-played \footnote{The threshold chosen is to be one quarter of the numerical range $[-1,1]$, i.e.\,$-0.5$. Keys who admit values below, above are taken as non-played and played, respectively.}.
The results of the TFC-D1 for a selection of test set time-series  are depicted in fig.\,\eqref{fig:results_JSB}. 
\begin{figure}[!ht]
     \centering \vspace{-0,1 cm}
\begin{tabular}{SSSSS}  \toprule
    {$\text{networks}$} & {$\text{weights}$} &  {$\stackrel{\text{one-step}}{\text{mse  }}$} &  {$\stackrel{\text{one-step }}{\text{accuracy }}$} & {$\stackrel{\text{CIFAR 10}}{\text{accuracy }} $} \vspace{0.2 cm} \\ \midrule
    {$\text{LSTM}$}  & 2.782M & 0.1225 & 45.60 {$\%$ }&  \textemdash  \vspace{0.04cm} \\\midrule
     {$\text{TFC-D1}$}  & 1.270M  & 0.1078 & 55.03 {$\%$} &  \textemdash \vspace{0.04cm}	\\
   { $\text{TFC-D1$^\circ$}$}  & 1.271M  & \textemdash &  \textemdash  & 90.42 {$\%$ } \\ \bottomrule
\end{tabular}
 \caption{Performance comparison in between the LSTM and the $D=1$ TFC network  denoted as TFC-D1.  Moreover, the performance of the TFC-D1$^\circ$  network on the CIFAR 10 image classification task.}  \label{fig:results_JSB_table}
\end{figure} \vspace{-0,1 cm}
    
\textbf{Generalizability of the TFC architecture.}
%%%%%%%%%%%%%%%%%%%%%%%%%%%%%%%%%%%%%%%
%\subsection{ Generalizability of the TFC architecture }\label{sec:JSB2}
%%%%%%%%%%%%%%%%%%%%%%%%%%%%%%%%%%%%%%%
It is of great  interest to understand the capability of neural networks to generalize to other tasks. The TFC architecture is  composited of a residual network and an incriminator cell. Residual networks were originally introduced for image classification tasks \cite{he2015deep}. It is thus not too surprising that one may use a TFC architecture with a minor change to make it applicable for those setups. 
With that in mind we modify the TFC-D1 denoted as TFC-D1$^\circ$ to make it applicable to the CIFAR 10 image classification task. The network architecture is modified by attaching a fully connected layer before the output to give ten categories and moreover change the objective function to be of categorical cross-entropy type. Secondly we change the hyper-parameter of the incriminator cell to $r = 0, t= 1, N = 1200$ and change the optimizer learning rate to $0.0005$. The training is over 450 epochs.  The accuracy of the TFC-D1$^\circ$ on the CIFAR 10 image classification task is $90.42\%$, see fig.\,\eqref{fig:results_JSB_table}. Concludingly, there is virtually no change to the  main architecture nor hyper-parameter optimization compared to the TFC-D1 which was used for the time series forecast of the JSB chorals. Thus we suggest that the TFC architecture admits potential to be generalized even further by e.g.\,advancing the notion and functionality of the incriminator cell. 

%%%%%%%%%%%%%%%%%%%%%%%%%%%%%%%%%%%%%%%
\section{ Conclusions}
%%%%%%%%%%%%%%%%%%%%%%%%%%%%%%%%%%%%%%%
In this work we proposed a novel architecture for time series forecasting using non-causal convolutional layers acting democratically on the spatial and time directions of the data. We empirically test  the TFC architecture on the sequential MNIST data set as well as JSB chorals dataset and we find that it outperforms conventional recurrent techniques. Furthermore, we show that the network can directly be applied to the different task of image classification of the CIFAR 10 data set.

As a future direction it would be interesting to draw the direct comparison to  other convolutional techniques for time series forecasting such as dilated convolutional nets. Secondly, one may benchmark TFC's on precipitation nowcasting.
Lastly, it would be of great interest to extend the TFC architecture to three dimensional time series data e.g.\,to sequences of three dimensional renderings. This however, requires four-dimensional convolutional layers. 

\vspace{1 cm}
\acknowledgements

Many thanks to Yoshinobu Kawahara (Kyushu University and AIP Center, RIKEN) for his support and useful comments on the draft. In particular, I want to take the opportunity to express my gratitude to the Kavli IPMU at Tokyo university where  part of this work was conducted. This work was supported by the WPI program of Japan.\vspace{1 cm}

%\begin{thebibliography}{15}
\nocite{*}
\bibliographystyle{utcaps}
%\newpage
\bibliography{references}

 \appendix
  
 \section{Layer Architecture and Training }\label{app:DetailsArch} %\vspace{-0.3cm}
 
The experiments of sections \ref{sec:MNIST} and \ref{sec:JSB} are performed on a single NVIDIA Tesla K80 GPU.  The networks are implemented in Keras for Tensorflow. The Tensorflow version is $1.15.0$. We choose notation similar to Keras' to express the pseudo code. Let us first set up some notation.
\newline \vspace{0.04 cm}
$  CL2D^{\tiny (n) , True}_{\tiny (r_1,r_2)}$ abbreviates a  two-dimensional  convolutional  LSTM layer with n features and kernel size of $(r_1,r_2)$ while the setting of the return sequence is true.
\newline \vspace{0.04 cm}
$  Conv2D^{\tiny (n) }_{s \,,\tiny (r_1,r_2)}$ is a two-dimensional convolutional  layer with n features and kernel size of $(r_1,r_2)$ and strides  s along one dedicated direction. The other strides direction is equal to one. Padding to equal size, i.e.\,"same" by default.
\newline \vspace{0.04 cm}
$ Conv3D^{\tiny (n) }_{s , \, \tiny (r_1,r_2,r_3)}$ is a  three-dimensional  convolutional  layer with n features and kernel size of $(r_1,r_2,r_3)$. All other strides are equal to one. Padding to equal size, i.e.\,"same" by default.
\newline \vspace{0.04 cm}
$  FC^{\tiny (n)}$ abbreviates a fully connected  layer with n-features.
\newline
$  CC$ denotes a concatenate layer.
\newline
\textbf{Residual Cell.} Let us next give the explicit definition of the residual cell used in the networks. In contrast to the notion introduced  around eq.\eqref{residual_Block} it admits multiple hyperparameters. We omit  dropout layers, and additional intermediate activation layers for simplicity. 
\newline
$R^{3D \,\,  (n_1;\, n_2)}_{ s_1,\,(r_1,r_2,r_3) \, ; \,\, s_2 \, (t_1,t_2,t_3)} $ is the three dimensional residual cell which admits  the following architecture
\bea\label{residual_Cell}
\text{Layer}_1 &=  Conv3D^{\tiny (n_2) }_{s_2 , \, \tiny (t_1,t_2,t_3)} \circ Conv3D^{\tiny (n_1) }_{s_1 , \, \tiny (r_1,r_2,r_3)}  \nonumber \\
\text{Layer}_2  &=  \begin{cases} 
      s_1 > 1\, \text{or} \; s_2 > 1 \;\; \; Conv3D^{\tiny (n_1/2) }_{s_1 + s_2 , \, \tiny (1,1,1)} \\
     	\text{else}  \;\; \;  \text{identity} \vspace{-0.08 cm}
   \end{cases} 
 \eea \vspace{-0.08 cm}
  \beq
R^{3D \,\,  (n_1;\, n_2)}_{ s_1,\,(r_1,r_2,r_3) \, ; \,\, s_2 \, (t_1,t_2,t_3)}  :=  CC\big(\text{Layer}_1 , \;  \text{Layer}_2 \big)\nonumber \;\; .
\eeq 
Let us next turn to the lower-dimensional case. \vspace{0.04 cm}
\newline
$R^{2D \,\,  (n_1;\, n_2)}_{ s_1,\,(r_1,r_2) \, ; \,\, s_2 \, (t_1,t_2)} $ is the two-dimensional residual cell  which admits  the analogous architecture to \eqref{residual_Cell} but with the $Conv3D$ layer substituted by a $Conv2D$ layer. Padding to equal size, i.e.\, "same" by default.
\newline
\textbf{Incrimator Cell.}  The incriminator cell makes use of the function \textit{ExtractImagePatches}  from the library \textit{tensorflow.python.ops.array\_ops} which we denote by $EP_{\tiny(r_1,r_2)}$ with kernel-size $(1,r_1,r_2,1)$, padding of  type "same" and strides as well as rates of $(1,1,1,1)$.  Please view the \href{https://www.tensorflow.org/api_docs/python/tf/image/extract_patches}{tensorflow documentation} for details. 
\newline
 $\mathcal{I}^{\tiny (2D)}_{r,t,N}$ is the two-dimensional incriminator cell defined as \vspace{- 0,2 cm}
 \beq\label{inc2d}
L_1( \{  x,  \, y \} ) =  CC\big( EP_{\tiny(r,r)}  ( x ) ,\;  EP_{\tiny(t,t)} (  y )\big)  \vspace{- 0,04 cm} \;\; ,
\eeq
\beq
  \mathcal{I}^{\tiny (2D)}_{r,t,N} \big( \{  x^{\tiny(m)}  ,  \, y^{\tiny(n)}\}\big) =  FC^{\tiny(m)} \circ  FC^{\tiny(N)}  \circ L_1( \{  x^{\tiny(m)}  ,  \, y^{\tiny(n)}\} ) \; , \nonumber 
   \eeq
where $x, \, y$ is two-dimensional spatial data, respectively. 
\newline
$\mathcal{I}'^{\tiny (2D)}_{r,t,N}$ the primed incriminator cell is defined according to eq.\eqref{inc2d} modified as in eq.\eqref{incCelladvanced}.
\newline
$\mathcal{I}^{\tiny (1D)}_{r,t,N}$ and $\mathcal{I}'^{\tiny (1D)}_{r,t,N}$ are defined analogously as the two-dimensional case but with kernel of \textit{ExtractImagePatches}  as  $(1,1,r,1)$ thus with $EP_{\tiny(1,r)}$ and $EP_{\tiny(1,t)}$.

\section{Networks Seq. MNIST}\label{app:DetailsMNIST}%\vspace{-0.3cm}
The input to all networks is a sequence of ten single-band images of shape $64 \times 64$ pixels, i.e.\,of input shape $(10,64,64,1)$.  We rescale the data to take values in $[-1,1]$. 
 \begin{figure}[!ht]
     \centering\vspace{-0.3 cm}
\begin{tabular}{ccc} \toprule
    {$\text{Networks}$} & {$\text{Architecture}$}   \vspace{0.2cm}\\ 
    $\text{\small ConvLSTM}$  &
     \small{$CL2D^{\tiny (8) , True}_{\tiny (7,7)}, \, CL2D^{\tiny (16) , True}_{\tiny (5,5)}, $ } \vspace{0.1cm}    \\
     &   \small{$CL2D^{\tiny (32) , False}_{\tiny (3,3)} ,  \, FC^{\tiny(32)} , \, FC^{\tiny (1)} $ }     \vspace{0.1cm}  \vspace{0.3cm}  \\   
    $\text{\small ConvLSTM-L}$  &     \small{$CL2D^{\tiny (16) , True}_{\tiny (7,7)},  \, CL2D^{\tiny (32) , True}_{\tiny (5,5)}, $} \, \vspace{0.1cm}  \\
    & \small{$CL2D^{\tiny (64) , False}_{\tiny (3,3)} ,  \, FC^{\tiny(64)} , \, FC^{\tiny(1)}   $ }   \vspace{0.1cm}   \vspace{0.3cm} \\ 
    $\text{\small TFC-D2}$  &    \small{$R^{3D \,\,  (16;\, 32)}_{ 1,\,(4,8,8) \, ; \,\,1 \, (4,5,5)} , \, R^{3D \,\,  (32;\, 64)}_{ 2,\,(2,5,5) \, ; \,\,2 \, (4,3,3)},  $}\, \vspace{0.1cm} \\
    &\small{$ R^{3D \,\,  (72;\, 96)}_{ 2,\,(3,2,2) \, ; \,\,2 \, (2,3,3)}  , \; FC^{\tiny(96)} , \, FC^{\tiny(8)}, \; \mathcal{I}_{7,1, 1000} $ }  \vspace{0.1cm}  \vspace{0.3cm}  \\    
    $\text{\small TFC-D2'}$  &    \small{$R^{3D \,\,  (16;\, 32)}_{ 1,\,(4,8,8) \, ; \,\,1 \, (4,5,5)} , \, R^{3D \,\,  (32;\, 64)}_{ 2,\,(2,5,5) \, ; \,\,2 \, (4,3,3)},  $}\, \vspace{0.1cm} \\
    &\small{$ R^{3D \,\,  (72;\, 96)}_{ 2,\,(3,2,2) \, ; \,\,2 \, (2,3,3)}  , \; FC^{\tiny(96)} , \, FC^{\tiny(8)}, \; \mathcal{I}'_{7,1,1,1000} $ }  \vspace{0.1cm}  \vspace{0.3cm}  \\    
    $\text{\small TFC-D2-L}$  &    \small{$R^{3D \,\,  (16;\, 32)}_{ 1,\,(4,8,8) \, ; \,\,1 \, (4,5,5)} , \, R^{3D \,\,  (32;\, 64)}_{ 1,\,(2,5,5) \, ; \,\,2 \, (4,3,3)},   $}\, \vspace{0.1cm} \\ 
      &\small{$ R^{3D \,\,  (64;\, 64)}_{ 1,\,(2,5,5) \, ; \,\,2 \, (4,3,3)},        \,  R^{3D \,\,  (72;\, 72)}_{ 1,\,(3,2,2) \, ; \,\,2 \, (2,3,3)} , $} \vspace{0.1cm}   \\
    &\small{$ R^{3D \,\,  (72;\, 128)}_{ 1,\,(3,2,2) \, ; \,\,2 \, (2,3,3)}  , \; FC^{\tiny(128)} , \, FC^{\tiny(8)}, \; \mathcal{I}_{7,1,1000} $ }  \vspace{0.1cm}  \vspace{0.3cm}  \\    

    $\text{\small TFC-D2-L'}$   &    \small{$R^{3D \,\,  (16;\, 32)}_{ 1,\,(4,8,8) \, ; \,\,1 \, (4,5,5)} , \, R^{3D \,\,  (32;\, 64)}_{ 1,\,(2,5,5) \, ; \,\,2 \, (4,3,3)},   $}\, \vspace{0.1cm} \\ 
      &\small{$ R^{3D \,\,  (64;\, 64)}_{ 1,\,(2,5,5) \, ; \,\,2 \, (4,3,3)},        \, R^{3D \,\,  (72;\, 72)}_{ 1,\,(3,2,2) \, ; \,\,2 \, (2,3,3)} , $} \vspace{0.1cm}   \\
    &\small{$ R^{3D \,\,  (72;\, 96)}_{ 1,\,(3,2,2) \, ; \,\,2 \, (2,3,3)}  , \; FC^{\tiny(96)} , \, FC^{\tiny(8)}, \; \mathcal{I}'_{7,1,1,1000} $ }  \vspace{0.1cm}  \vspace{0.3cm}  \\    
\end{tabular} \vspace{-0.6 cm}
 \caption{Details of the networks architectures.  We omit  dropout layers,  batch-normalization, permutation and additional intermediate activation layers for simplicity. The final activation layer is $tanh$.
}   \label{fig:results_MNIST_app}
\end{figure}
As for the TFC networks in our notation for the residual block one may write e.g.\, 
\beq
\text{TFC-D2} \;  =  \; \mathcal{I}_{7,1, 1000}   \circ  \mathcal{R}_8   \;\; ,
\eeq
 Note that all networks have eight features in their residual block, see fig.\eqref{fig:results_MNIST_app}. Also  let us point out that the  $L$ -extension  networks in fig.\eqref{fig:results_MNIST_app} have  two residual cells added in the middle and a change in the strides. The hyperparameters for the kernel sizes are the same.

\textbf{Strategy hyperparameters.} Let us  next comment on choosing the hyperparameters for the residual block, in particular, the kernel sizes of the convolution operations. The strategy we followed is to make the spatial kernel size larger than the time kernel direction. This is motivated by the analogy to the speed of data moving across time-steps, i.e.\,spatial kernel over time kernel size. Thus larger ratios will capture more effects while keeping the time kernel number small. However, one also may want to increase the size of the time-direction kernel to capture correlations of the data across several time steps. 
Those are the two main guiding principles  towards optimizing  the residual block in TFC networks.
\newline
  \vspace{0,5 cm}
 \section{Networks JSB Chorals}\label{app:JSBDetails}%\vspace{-0.3cm}
 The input to all networks is a sequence of ten single-band images of shape $1 \times 88$ pixels, i.e.\,of input shape $(10,88,1)$ for the JSB chorals dataset. 
 For the CIFAR 10 dataset the images are colored and of shape $32 \times 32$ pixels, i.e. of input shape $(32,32,3)$.
 We rescale the data to take values in $[-1,1]$. 
One may write  the TFC network as
\beq \text{TFC-D1} \; =  \;   \mathcal{I}_{12,1,1200}  \circ \mathcal{R}_{16} \;\; .
\eeq
 Let us close by emphasizing that the networks TFC-D1 and TFC-D1$^\circ$ admit the same main components and hyperparameters fig.\eqref{fig:results_JSB_table_app}.  
 \begin{figure}[!ht]
     \centering  \vspace{-0.1 cm}
\begin{tabular}{SS}  \toprule
    {$\text{Networks}$} & {$\text{Layers}$} \vspace{0.2 cm} \\ \midrule
    {$\text{LSTM}$}  &       \small{$ LSTM_{128}^{true} ,  \; LSTM_{256}^{true} ,\;  TD_{256} \;, LSTM_{256}^{true} ,$}\, \vspace{0.1cm} \\ 
      &\small{$  TD_{256} \;, LSTM_{512}^{true}\;, FC^{88} $} \vspace{0.1cm}  
\vspace{0.3cm}  \\    
     {$\text{TFC-D1}$} &    \small{$R^{2D \,\,  (16;\, 32)}_{ 1,\,(5,16) \, ; \,\,1 \, (6,15)} , \, R^{2D \,\,  (32;\, 64)}_{ 1,\,(6,8) \, ; \,\,1 \, (3,4)},   $}\, \vspace{0.1cm} \\ 
      &\small{$ R^{2D \,\,  (64;\, 72)}_{ 1,\,(3,5) \, ; \,\,2 \, (4,7)},        \, R^{2D \,\,  (72;\, 96)}_{ 1,\,(4,8) \, ; \,\,2 \, (4,7)} , $} \vspace{0.1cm}   \\
    &\small{$ R^{2D \,\,  (96\, 96)}_{ 1,\,(2,8) \, ; \,\,2 \, (3,7)}  ,   R^{2D \,\,  (96;\, 128)}_{ 1,\,(3,2) \, ; \,\,2 \, (2,3)} $ }  \vspace{0.1cm}  \\
    &\small{$ FC^{\tiny(128)} , \, FC^{\tiny(16)}, \; \mathcal{I}_{12,1,1200} $ }  \vspace{0.1cm}  \vspace{0.3cm}  \\    
   { $\text{TFC-D1$^\circ$}$}  &    \small{$ \, \text{TFC-D1}, \,  Flatten , \, FC^{\tiny(96)} , \, FC^{\tiny(10)}  $} \\ \bottomrule
\end{tabular} 
 \caption{Details of the networks architectures. We omit  dropout layers,  batch-normalization, permutation and additional intermediate activation layers for simplicity. Let us just stress that we use the final softmax activation for TFC-D1$^\circ$.  $LSTM_n^{true}$ is a LSTM  cell  with n-features  with return sequence true.  $TD_n$ is a dime-distributed fully connected layer with n-features. }  \label{fig:results_JSB_table_app}
\end{figure}  \vspace{-0.5 cm}

%\end{thebibliography}
\end{document}